\documentclass[letterpaper]{article} 
\usepackage{aaai2027}

\usepackage[hyphens]{url}  
\usepackage{graphicx} 
\usepackage{natbib}  
\usepackage{caption} 
\usepackage{algorithm}
\usepackage{algorithmic}
\usepackage{amsmath, amssymb}
\usepackage{multirow}
\usepackage{multicol}
\usepackage{colortbl}

\usepackage{url}
\usepackage{hyperref}

\usepackage{newfloat}
\usepackage{listings}
\DeclareCaptionStyle{ruled}{labelfont=normalfont,labelsep=colon,strut=off} 
\floatstyle{ruled}
\newfloat{listing}{tb}{lst}{}
\floatname{listing}{Listing}

\usepackage{booktabs}
\usepackage{amsmath}
\title{ Practice Makes Policies: Bootstrapping and Consolidating Robotic Capabilities from Zero Human Demonstrations}

\author {
    Jialiang Li\textsuperscript{\rm 1},
    Yuhan Wang\textsuperscript{\rm 1},
    Haojun Li\textsuperscript{\rm 1},
    Gaojing Zhang\textsuperscript{\rm 2}, 
    Yangtian Ye\textsuperscript{\rm 1}, \\
    Qipeng Liu\textsuperscript{\rm 1},
    Haotian Liang\textsuperscript{\rm 1},
    Wenzhao Lian\textsuperscript{\rm 1},
}
\affiliations {
    \textsuperscript{\rm 1}Shanghai Jiao Tong University\\
    \textsuperscript{\rm 2}University of Sussex\\
    lijl25@sjtu.edu.cn
}

\begin{document}

\makeatletter
\let\@oldmaketitle\@maketitle
\renewcommand{\@maketitle}{\@oldmaketitle
  \begin{center}
    \includegraphics[ width=0.95\textwidth]{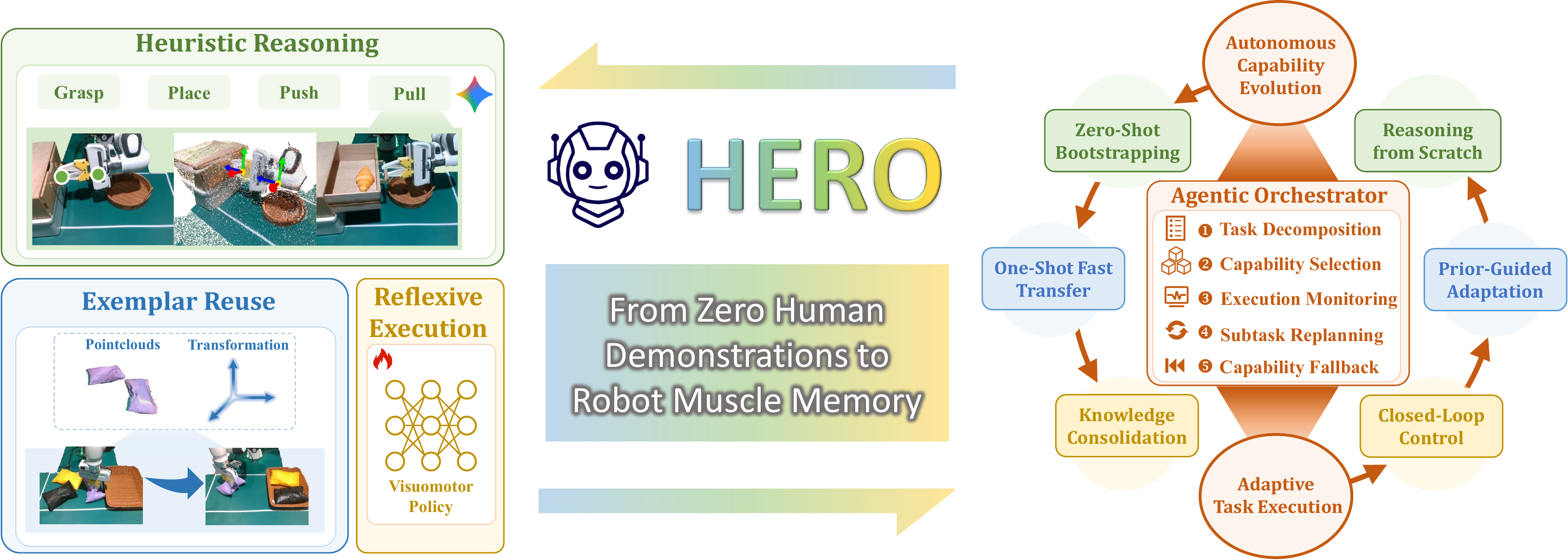}
  \end{center}
  \refstepcounter{figure}
  \normalsize{Figure~\thefigure: Overview of the HERO framework. HERO enables continuous capability evolution from zero human demonstration to robot muscle memory. (1) Left: HERO incorporates three execution capabilities:  Heuristic Reasoning for zero-shot bootstrapping,  Exemplar Reuse for one-shot fast transfer, and Reflexive Execution for robust visuomotor control. (2) Right: Centered around an agentic orchestrator, HERO seamlessly coordinates two operational loops: autonomous data evolution for experience accumulation and adaptive task execution for reliable real-world deployment. }%
  \label{fig:teaser}
  \medskip
}
\makeatother

\maketitle

\begin{abstract}
General-purpose robotic manipulation requires robots to perform diverse tasks in open-world environments while improving their skills over time. Despite recent progress in robotic manipulation, existing systems still primarily acquire manipulation skills in a static manner, where capabilities are learned for specific tasks or settings rather than adaptively evolving through physical interaction. Resembling how repeated practice enables humans to develop muscle memory, advanced manipulation proficiency requires an autonomous capability evolution mechanism that allows robots to progressively transform interaction experiences into increasingly effective manipulation abilities. To this end, we propose HERO, a self-improving hierarchical embodied agent that enables autonomous capability evolution from zero human demonstrations. HERO organizes heuristic reasoning, exemplar reuse, and reflexive execution into a unified orchestration framework, allowing robots to autonomously bootstrap manipulation experience, rapidly accumulate reusable behaviors through experience transfer, and progressively consolidate recurring interactions into efficient closed-loop visuomotor policies. By tightly coupling autonomous data collection with task execution, HERO continuously expands and dynamically schedules manipulation capabilities according to different stages of experience accumulation and execution requirements. Extensive experiments demonstrate that HERO substantially reduces human intervention during robotic data collection while achieving robust manipulation across diverse tasks, providing a promising path toward self-improving robotic systems. View our project in \href{https://hero-agent.github.io/}{https://hero-agent.github.io/} .

\end{abstract}

\section{Introduction}

General-purpose robots are expected to execute diverse long-horizon manipulation tasks in open-world environments with minimal human supervision. Achieving this goal requires not only solving individual manipulation tasks, but also continuously acquiring, reusing, and improving robotic capabilities through interaction with the physical world. 

Existing approaches to open-world manipulation typically face a fundamental trade-off between zero-shot generalization and efficient closed-loop execution. Modular agentic frameworks \cite{CodeAsPolicies, ASP,HoloAgent} leverage foundation models for general reasoning, but suffer from high inference latency and remain largely static due to limited experience reuse. End-to-end visuomotor policies \cite{RT-2,OpenVLA,Pi0.5} enable efficient reactive control, yet rely heavily on human demonstrations and lack autonomous capability acquisition. Recent self-improving embodied agents \cite{RoboClaw, ABotClaw, ASPIRE,ENPIRE} automate experience collection, but focus on task-specific optimization rather than a unified mechanism for transforming reasoning into increasingly efficient embodied behaviors. Therefore, a critical challenge remains: how can a robot bootstrap from general-purpose reasoning, autonomously accumulate physical experiences, and progressively compile them into efficient manipulation capabilities?

To address this challenge, as presented in Figure \ref{fig:teaser}, we propose \textbf{HERO}, a self-improving hierarchical agentic framework that unifies capability acquisition, evolution, and execution. Rather than relying on a monolithic policy, HERO organizes robotic intelligence into three capability levels: (1) \textbf{H}euristic reasoning, which leverages VLM-guided grounding to synthesize behaviors for novel tasks; (2) \textbf{E}xemplar reuse, which adapts previously acquired successful experiences through spatial motion transfer; and (3) \textbf{R}eflexive execution, which distills manipulation experiences into closed-loop visuomotor policies. These heterogeneous capabilities are coordinated by a hierarchical \textbf{O}rchestrator, which manages both capability evolution and execution by determining how experiences are acquired, reused and consolidated.

Starting from zero human demonstrations, HERO establishes an autonomous capability evolution process that progressively transforms general reasoning into embodied capabilities. During data collection, the orchestrator selects appropriate acquisition strategies, leveraging heuristic reasoning for unseen scenarios and exemplar reuse when transferable experiences are available for fast adaptation. Successful executions are automatically evaluated, where frequently encountered behaviors are consolidated into reflexive policies. 

During deployment, HERO dynamically schedules capabilities based on experience availability and execution feedback. Reflexive execution are utilized for established behaviors, while exemplar reuse and heuristic reasoning provide alternative mechanisms when direct policy execution is insufficient. This dynamic capability scheduling allows HERO to effectively exploit accumulated experiences while maintaining flexibility across diverse manipulation conditions.

Extensive experiments demonstrate that HERO enables autonomous capability evolution with minimal human intervention and achieves robust manipulation through adaptive execution across diverse tasks.

In summary, we make the following contributions:

\begin{itemize}
     \item \textbf{Hierarchical Capability Orchestration}. We introduce HERO, a self-improving embodied agent that unifies heuristic reasoning, exemplar reuse, and reflexive visuomotor execution for open-world manipulation.
     \item \textbf{Autonomous Capability Evolution}. We establish a capability evolution paradigm that bootstraps manipulation from zero human demonstrations and transforms physical experiences into reusable capabilities.
     \item \textbf{Adaptive Task Execution}. We develop an adaptive task execution mechanism that dynamically exploits different capabilities during execution, balancing efficiency and robustness in manipulation.
\end{itemize}


\section{Related Work}

\subsection{Embodied Agentic Orchestration}
Recent works formulate robot manipulation as an agentic orchestration problem by coordinating heterogeneous capabilities. Early language-guided frameworks ground high-level reasoning into executable robot skills \cite{SayCan,ProgPrompt,InnerMonologue,VoxPoser}, followed by modular systems that decompose instructions into executable subtasks through structured perception and planning modules \cite{ASP,HoloAgent}. Programming agents further realize such compositionality through executable code generation \cite{CodeAsPolicies,CapX}. Although these approaches improve task generalization, they often trade off broad capability coverage with reliable physical execution. Some physical orchestrators integrate VLA policies into modular workflows through monitoring and recovery \cite{VoloAgent,HAMSTER,HiRobot,FailSafe}, but primarily coordinate static capabilities without enabling their evolution. HERO instead enables reasoning, experience reuse, and reflexive visuomotor policies to progressively evolve from acquired experiences and be dynamically coordinated within a unified framework.

\subsection{Embodied Agentic Self-Improvement}
Recent works investigate scalable data generation and lifelong learning with reduced human supervision \cite{GenieCenturion,FieldGen, Act2Goal}. RoboClaw \cite{RoboClaw} further enables self-resetting data collection for continual policy refinement, but still relies on initial human demonstrations and frequent early-stage interventions. Inspired by open-ended skill acquisition and automated optimization \cite{Voyager,Eureka,DrEureka}, recent embodied systems pursue continual capability evolution. ASPIRE \cite{ASPIRE} acquires reusable robotic knowledge from execution traces through evolving skill libraries, while largely relying on simulation-based verification and reset mechanisms. ENPIRE \cite{ENPIRE} further automates experiment design and policy optimization through agentic workflows. In contrast, HERO enables capability evolution directly within a execution loop, unifying autonomous bootstrapping, experience reuse, and progressive refinement of robotic skills.

\section{Method}


As illustrated in Figure ~\ref{fig:teaser}, we propose \textbf{HERO}, a hierarchical agentic framework for autonomous capability evolution and adaptive task execution with zero human demonstrations. Mirroring how human cognition progresses from  deliberate reasoning to  muscle memory, HERO models robot learning and execution as two complementary processes traversing a shared capability spectrum, spanning \textbf{H}euristic  reasoning, \textbf{E}xemplar reuse, and \textbf{R}eflexive execution. A closed-loop \textbf{O}rchestrator unifies these capabilities through planning, execution monitoring, and capability scheduling, enabling continuous capability evolution from heuristic reasoning to reflexive execution and dynamic transitions among different capabilities during task execution.

\subsection{HERO Capability Hierarchy}
\label{sec:hierarchy}
The HERO capability spectrum is realized through three execution layers, each representing a distinct level of robot capability: a heuristic bootstrapper (L1) for zero-shot reasoning, an exemplar accelerator (L2) for one-shot experience fast adaptation, and a reflexive policy (L3) for efficient closed-loop visuomotor control. We elaborate each layer below.

\subsubsection{Heuristic Bootstrapper (L1)}

\begin{figure}
    \centering
    \includegraphics[ width=1\linewidth]{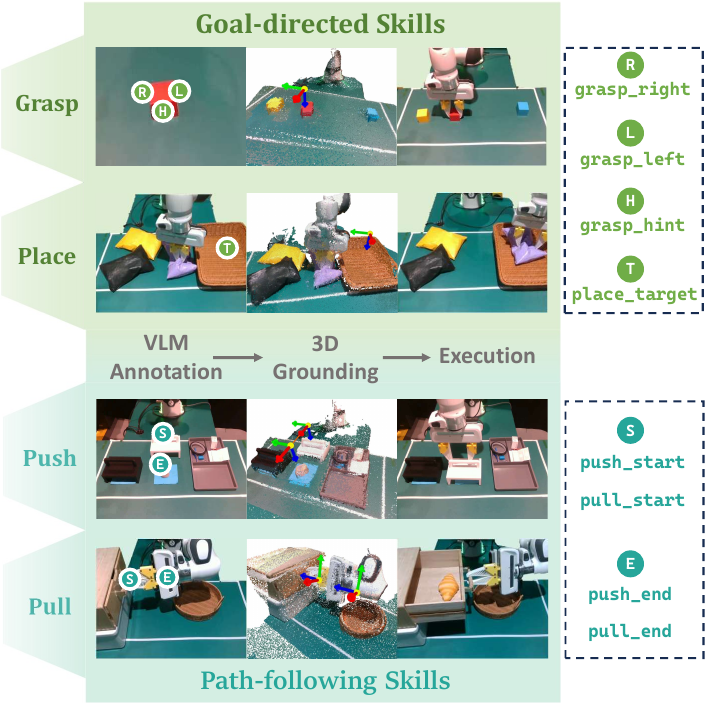}
    \caption{Visualization of the four primitive action skills in L1, illustrating the process from VLM annotation to 3D grounding and robot action execution.}
    \label{fig:skills}
\end{figure}

This layer operates as a zero-shot agentic module designed to execute tasks without requiring any prior human demonstrations. Given a language instruction and the visual observations, a VLM decomposes the semantic objective into a sequence of predefined primitive action skills. To enable geometric grounding, we reconstruct the 3D workspace from multi-view RGB observations and calibrated camera extrinsics using Depth-Anything-3~\cite{DA3}. Each primitive skill is then instantiated upon the reconstructed workspace to produce executable  actions.  

As illustrated in Figure \ref{fig:skills}, the primitive skills are categorized into two groups according to the geometric structures.

\textbf{Goal-directed Skills} predict a target end-effector (EE) pose, denoted as  $\mathcal{G}=\{t_{ee},R_{ee}\}$, where $t_{ee} \in \mathbb{R}^3$ and $R_{ee} = [x_{ee}, y_{ee}, z_{ee}]  \in SO(3)$ are the target translation and orientation, respectively. We instantiate this representation with the two skills below.

\begin{itemize}
    \item \textbf{Grasp}. The VLM predicts three semantic 2D keypoints on the target object surface: $p_l$ and $p_r$ representing the gripper jaw boundaries, and $p_h$ specifying the x-axis alignment hint. These points are lifted into 3D space as $P_l, P_r, P_h \in \mathbb{R}^3$. The target EE pose is computed as:
        \begin{equation}
        \begin{aligned}
            t_{ee} &= \frac{P_l + P_r}{2}, & y_{ee} &= \frac{P_r - P_l}{\|P_r - P_l\|} \\
            z_{ee} &= \frac{(P_h - t_{ee}) \times y_{ee}}{\|(P_h - t_{ee}) \times y_{ee}\|}, & x_{ee} &= y_{ee} \times z_{ee}
        \end{aligned}
        \end{equation}
    To resolve potential axis inversions caused by relative camera-robot frame configuration, we automatically rectify inconsistent axis directions to maintain workspace consistency. The resulting pose is executed by moving the EE to the predicted pose and closing the gripper.
    
    \item \textbf{Place}. The VLM predicts a single semantic 2D keypoint indicating the desired placement location, which is lifted into the reconstructed 3D workspace to obtain the target translation $t_{ee}$. The EE orientation $R_{ee}$ is inherited from the preceding manipulation state, and the resulting pose is executed by opening the gripper at the target location.
\end{itemize}

\textbf{Path-following Skills} predict a Cartesian motion segment, denoted as $\mathcal{T} = \{P_s, P_e, R_{ee} \}$, where $P_s$ and $P_e$ are the start and end positions of the motion, and $R_{ee}$ specifies the EE orientation maintained throughout the path. This is instantiated with the following two skills.

\begin{itemize}
    \item \textbf{Push}. The VLM predicts two semantic 2D keypoints corresponding to the push start and push end locations, which are lifted into the reconstructed 3D workspace as $P_s$ and $P_e$. The EE rotation $R_{ee}$ is determined by aligning (or anti-aligning) the TCP x-axis with the pushing direction $P_e - P_s$, maximizing the contact area between the gripper surface and the object for stable force transmission. 
    
    \item \textbf{Pull}. Similar to pushing, the VLM predicts the pull start and pull end keypoints to ground $P_s$ and $P_e$. Since pulling typically occurs when the robot already maintains a hold on an object interface, the execution orientation $R_{ee}$ is directly inherited from the preceding manipulation pose.
\end{itemize}

To compose individual primitives into executable manipulation sequences, we additionally apply predefined transition constraints to handle skill-to-skill motion continuity and contact safety. Detailed implementation is provided in the supplementary material.



\subsubsection{Exemplar Accelerator (L2)}

This layer operates as a one-shot module designed to accelerate execution through rapid demonstration adaptation. Based on MT3 \cite{MT3}, L2 estimates a rigid transformation between the current object and a retrieved exemplar using object pointclouds lifted from SAM3 \cite{SAM3} masks. This warps the entire reference trajectory to fit the new scene layout without multiple time-consuming VLM calls required by L1.

\textbf{Demonstration Retrieval}. Given the target language instruction, L2 queries an aggregated exemplar library containing historical successful trajectories and valid object pointclouds. We employ word-level LCS \cite{LCS} to identify similarities between the current instruction and past ones. A reference demonstration is deemed retrieved if its normalized similarity exceeds a predefined threshold.

\textbf{Motion Transfer}. Upon retrieval, L2 computes a 3D rigid transformation from coarse to fine. The translation is first derived from the centroid offset between the demonstration and current object pointclouds, while the initial rotation matrix is generated by an lightweight, pretrained PointNet++ \cite{PointNet++} regressor from MT3 \cite{MT3}. This transformation is then refined via  GICP \cite{GICP} and applied to warp the entire demonstrated sequence to yield an executable trajectory for the current scene.

\subsubsection{Reflexive Policy (L3)}

This layer operates as a learned, reactive control module designed to provide closed-loop motor control. Unlike L1 and L2, which generate executable trajectories based on the observations before execution, L3 directly maps language instructions and visual observations to robot actions through a visuomotor policy, continuously adapting motions based on real-time feedback. We instantiate L3 using $\pi_{0.5}$ \cite{Pi0.5}.

\subsection{HERO Agentic Orchestration}
\label{sec:orchestrator}

The three capability layers introduced above represent different ways of transforming and utilizing manipulation experiences. Through VLM-based reasoning, the HERO agentic orchestrator unifies these heterogeneous capabilities into an autonomous system by coordinating subtask planning, capability selection, execution monitoring,  and plan refinement.

Given a user-specified task and the current visual observations, the orchestrator decomposes the task into a dynamic subtask plan $\mathcal{P} = \{s_1, s_2, \dots, s_n\}$. Unlike approaches that rely on predefined templates \cite{RoboClaw}, the orchestrator generates subtasks directly in free-form natural language, allowing subtasks to be naturally described according to their objectives and contexts. The orchestrator then dispatches each subtask to an appropriate capability layer and continuously updates the plan based on execution feedback.

After each subtask execution step, the orchestrator compares the updated observations against the intended objective and determines the execution outcome as one of $ \{\text{\texttt{successful}}, \text{\texttt{failed}},  \text{\texttt{call\_human}}\}$. The execution outcome is then used to refine the subtask plan, including advancing completed subtasks, revising future subtask sequences, or requesting human intervention when necessary.

Built upon this orchestration framework, HERO supports two operational processes, as described below.

\subsubsection{H2E2R Autonomous Capability Evolution}
\label{sec:h2e2r}
Unlike conventional imitation learning pipelines that rely on manually collected demonstrations, the orchestrator follows a H$\rightarrow$E$\rightarrow$R capability progression to autonomously evolve robotic capabilities from zero human demonstrations, transforming heuristic reasoning into reusable experiences and finally consolidating them into visuomotor policies.

For an unseen subtask, the orchestrator first invokes L1, which performs VLM-guided reasoning to generate executable manipulation behaviors without requiring human demonstrations. Successful executions are evaluated and stored as reusable exemplars. As transferable experiences become available, the orchestrator dispatches corresponding subtasks to L2, which accelerates execution by adapting previously successful trajectories through fast motion transfer.

To enable long-term autonomous data collection with minimal manual intervention, the orchestrator synthesizes a reverse task after each completed task to restore the scene for the next collection cycle. The reverse task is inferred from the completed task, its historical subtask plan, and the current visual observations, before being decomposed and executed through the same orchestration framework. This design enables autonomous scene reset and supports continuous experience acquisition across repeated collection cycles.

Once sufficient experiences have been accumulated, the collected trajectories are filtered to remove idle frames and converted into training data for L3. The L3 policy is thus trained, consolidating previously acquired manipulation experiences into direct visuomotor control, completing the capability evolution from heuristic reasoning to exemplar reuse and finally reflexive execution.

\subsubsection{R2E2H Adaptive Task Execution}
\label{sec:r2e2h}
During task execution, HERO adopts the R$\rightarrow$E$\rightarrow$H capability progression: prioritizing closed-loop reflexive policies for established behaviors, leveraging exemplar reuse through prior successful trajectories, and invoking heuristic reasoning when needed.

The orchestrator first dispatches the subtask to L3 if it is available (i.e., a trained policy is applicable), L3 executes the assigned subtask until a termination condition is satisfied, determined either by action convergence over consecutive steps or by reaching a predefined maximum inference rounds. If the required policy is unavailable or repeated execution failures are detected, the orchestrator falls back to L2. Should no transferable exemplar be available or repeated adaptation failures occur, the subtask is finally dispatched to L1.  This adaptive capability scheduling allows HERO to effectively exploit accumulated experiences while retaining the flexibility to handle varying task conditions, enabling robust manipulation across diverse scenarios.

\section{Experiments}

In this section, we conduct a series of experiments to evaluate the effectiveness of our proposed method, and answer the following questions:

\textbf{Q1: } Can HERO autonomously collect effective experiences for capability evolution?

\textbf{Q2: } Does HERO’s orchestration effectively leverage different capabilities for adaptive task execution?

\textbf{Q3: } What are the runtime behaviors, efficiency characteristics and failure modes of HERO?

\subsection{Implementation Details}

\subsubsection{Task Setup} 
\begin{figure}[t]
    \centering
    \includegraphics[width=1.0\linewidth]{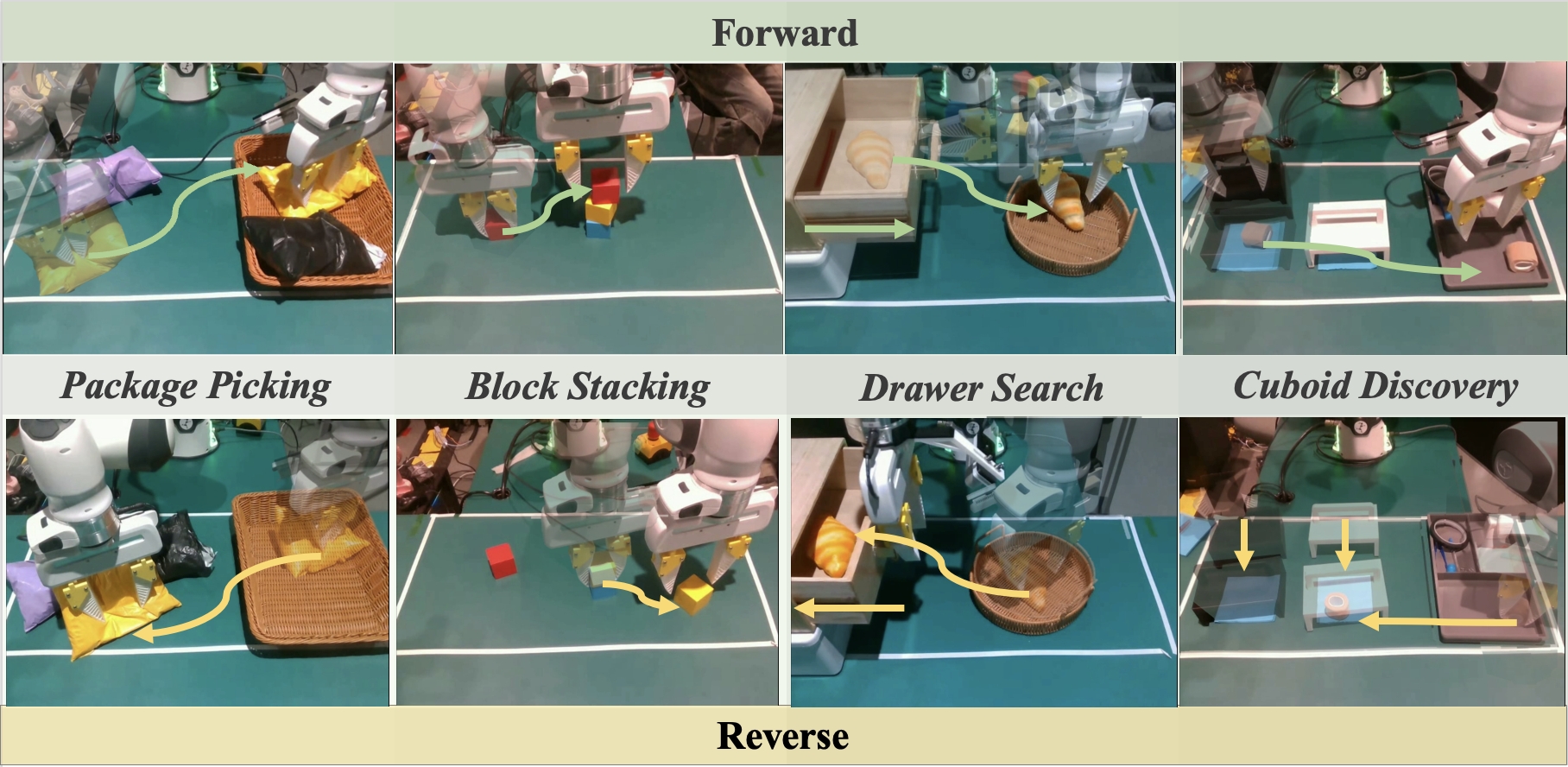}
    \caption{Illustration of the real‑world task settings. All tasks share basic manipulation skills (grasp and place). While the Drawer Search task additionally requires both pull and push, and the Cuboid Discovery task requires push only.}
    \label{fig:task-setup}
\end{figure}

As shown in Figure \ref{fig:task-setup}, We evaluate our framework across four distinct tasks in real-world environment, covering varied manipulation skills and semantic reasoning.  

\begin{itemize}
    \item \textbf{Package Picking }. Pick up packages with different colors and place them into a basket  (\textit{reverse task}: return packages to the table).
    \item \textbf{Block Stacking}. Stack blocks according to a specified top-down color sequence (\textit{reverse task}: move the stacked blocks back to the table). 
    \item \textbf{Drawer Search}. Open a drawer, retrieve the hidden croissant, and place it onto a plate  (\textit{reverse task}: return the croissant and close the drawer).
    \item \textbf{Cuboid Discovery }. Move occluding cuboids to discover a hidden bandage roll and place it on a gray tray (\textit{reverse task}: place the bandage roll under a random cuboid and recreate an occluded configuration).
\end{itemize}

\subsubsection{Hardware Setup} We employ a Franka Emika Panda robot arm and four RealSense D435i cameras for manipulation. Three exterior cameras provide RGB observations for L1/L2 scene reconstruction, while a front camera and a wrist-mounted one are used for L3 training and inference.

\subsubsection{Data Collection and Training} Our framework collects each task $30$ successful trajectories at $15$ Hz, in total $664$ of forward and reverse subtask episodes. For each task, we train a L3 policy on the corresponding dataset for $10000$ steps with batch size of $32$. Each task takes training on $2$ NVIDIA A100-SXM4-80GB GPUs for 5 hours.

\subsubsection{Execution Setup} We employ Gemini-3-Flash \cite{gemini} for all VLM queries. Each layer allows at most $3$ consecutive retries before fallback.  In L2, the LCS-based instruction retrieval filters out stop-words (e.g., ``the'', ``an'') and the similarity threshold is set to $1.0$. For L3, execution terminates upon action convergence (maximum consecutive joint difference $<0.02$ over the latest $8$ chunks) or after $25$ inference rounds.

\subsection{Autonomous Capability Evolution}

\begin{figure}[h]
    \centering
    \includegraphics[width=1\linewidth]{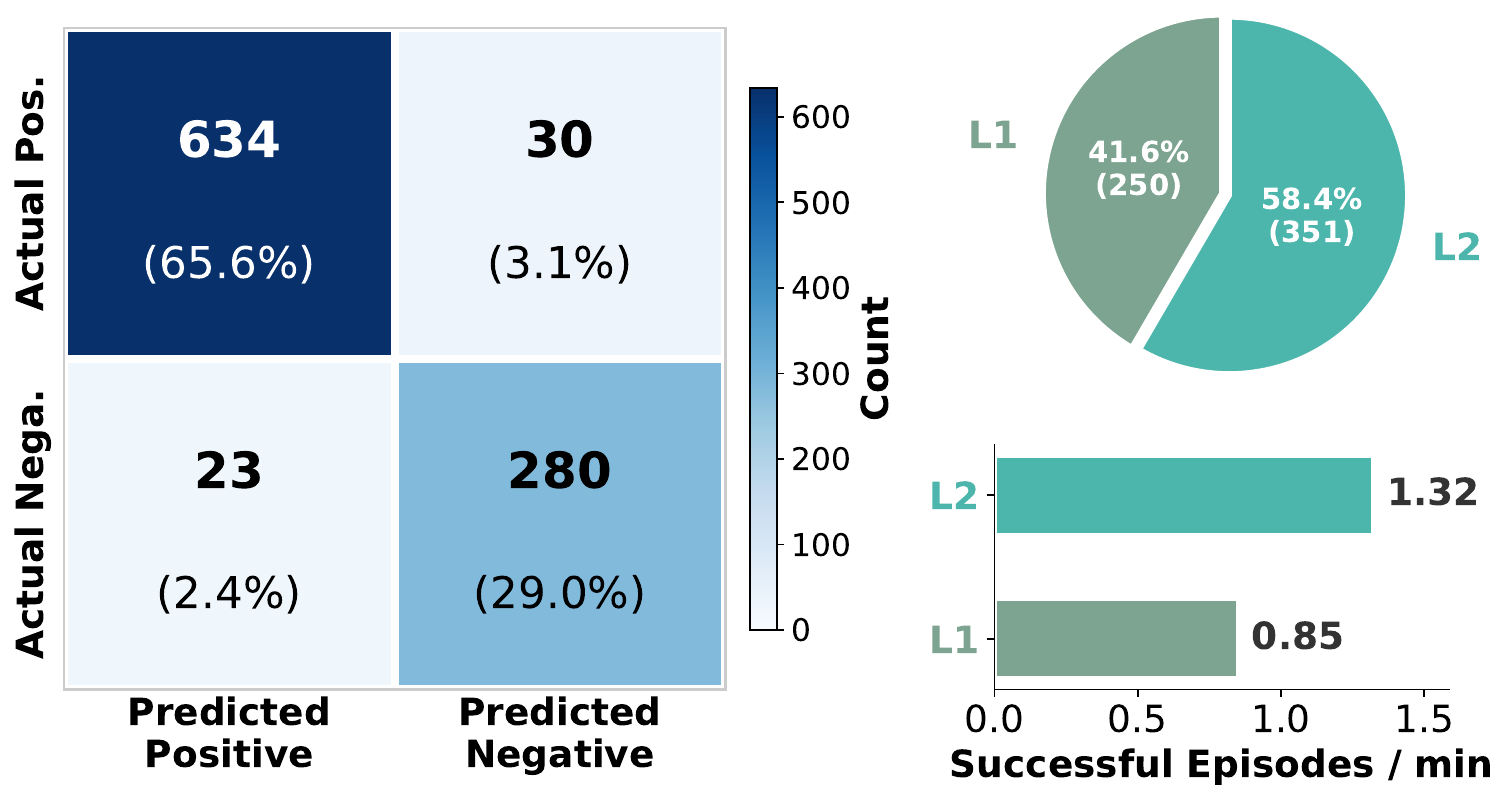}
    \caption{ (1) Left:  Subtask outcome prediction matrix demonstrating the orchestrator's reliable closed-loop monitoring. (2) Top Right: Data contribution ratio between L1 and L2. (3) Bottom Right: Data collection efficiency measured by successful subtask episodes collected per minute.}
    \label{fig:matrix}
\end{figure}

We answer \textbf{Q1} in this section by evaluating HERO's ability to autonomously acquire, curate, and transform physical experiences into evolving manipulation capabilities.

\textbf{Autonomous Data Collection.} Through the proposed H$\rightarrow$E$\rightarrow$R capability progression, HERO enables long-term autonomous data collection with minimal human involvement. Across all tasks, HERO collects 664 successful forward and reverse subtask episodes, while requiring only an average of $1.03s$ human intervention time per subtask for scene recovery. It further completes up to 46 consecutive autonomous subtask execution cycles without any human intervention. Among cases requiring reverse execution, reverse tasks achieve an average success rate of 84.3\%, enabling effective scene reset before requesting assistance.

 \textbf{Experience Curation.} Beyond autonomous data collection, HERO also automatically evaluates and curates collected experiences through the orchestrator. The left part of Figure \ref{fig:matrix} evaluates the reliability of the orchestrator’s data assessment by comparing its decisions with the actual execution outcomes. Across all tasks, 94.5\% of trajectories are correctly accepted after successful execution or rejected after failure, with only a small fraction requiring human correction. These results demonstrate that HERO can autonomously collect and curate high-fidelity manipulation experiences.

 \textbf{Experience Acquisition Analysis.} The right part of Figure \ref{fig:matrix} analyzes experience acquisition across capabilities. L1 and L2 provide comparable amounts of trajectory data (41.6\% and 58.4\%), where L1 expands coverage for unseen subtasks and L2 accelerates experience accumulation through exemplar reuse. Since execution success and collection speed exhibit different trade-offs, we evaluate data collection efficiency by measuring the number of successful subtask episodes collected per minute. L2 collects valid data faster than L1 ($1.32$ vs. $0.85$), demonstrating its advantage in accelerating data collection  through experience reuse.

\begin{figure}[h]
    \centering
    \includegraphics[width=0.8\linewidth]{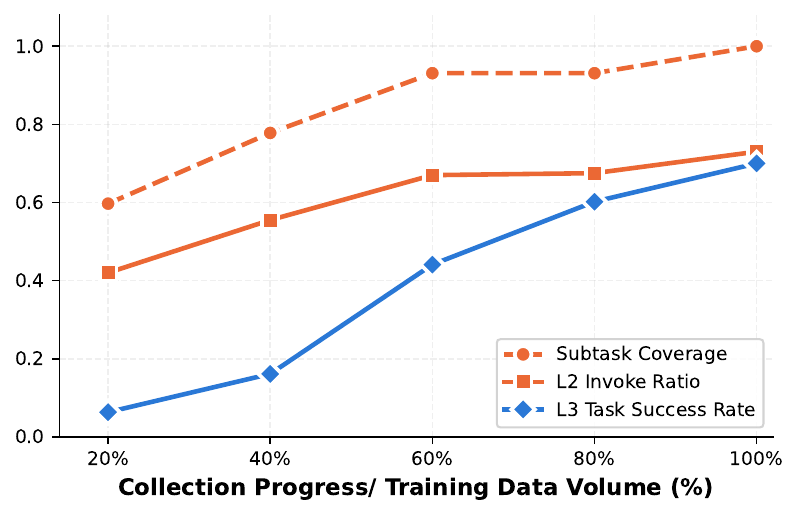}
    \caption{Capability evolution of HERO via autonomous experience accumulation.}
    \label{fig:evolution}
\end{figure}

\begin{table*}[]
    \centering
    \fontsize{9pt}{11pt}\selectfont
    
    \begin{tabular}{ c  c c  c c  c c  c c  c  c c  }
    \toprule
       \multirow{2}{*}{Method} & \multicolumn{2}{c}{Package Picking} & \multicolumn{2}{c}{Block Stacking} & \multicolumn{2}{c}{Drawer Search} & \multicolumn{2}{c}{Cuboid Discovery}  & \multicolumn{2}{c}{Mean} \\

        \cmidrule(lr){2-3} \cmidrule(lr){4-5}  \cmidrule(lr){6-7}  \cmidrule(lr){8-9}  \cmidrule(lr){10-11} 
         &  SR & TS(min) & SR & TS(min) & SR & TS(min) & SR & TS(min) & SR & TS(min)\\
         \midrule
        
        \textbf{HERO}  & \textbf{96.7\%} & 3.8 & \textbf{66.7\%} & 5.8 & \textbf{86.7\%} & 3.1 & \textbf{93.3\%} & 3.0 & \textbf{86.0\%} & 3.9 \\
        \midrule

        HERO (L1 only)  & 90.0\% & 4.6 & 60.0\% & 3.6 & 76.7\% & 3.5 & 76.7\% & 2.9 & 76.0\% & 3.7 \\
        HERO (L2 only)  & 86.7\% & 3.2 & 50.0\% & 3.1 & 63.3\% & 2.6 & 66.7\% & 2.1 & 66.7\% & 2.8 \\
        HERO (L3 only)  & 83.3\% & 3.8 & 56.7\% & 4.9 & 70.0\% & 2.7 & 70.0\% & 2.3 & 70.0\% & 3.4 \\
        HERO (L1 + L2)  & 76.7\% & 3.6 & 60.0\% & 6.3 & 73.3\% & 3.8 & 83.3\% & 2.1 & 73.3\% & 4.0 \\
        HERO (L1 + L3)  & 90.0\% & 4.3 & 66.7\% & 6.0 & 83.3\% & 3.9 & 90.0\% & 4.1 & 82.7\% & 4.6 \\
        HERO (L2 + L3)  & 86.7\% & 4.0 & 56.7\% & 5.3 & 70.0\% & 3.2 & 76.7\% & 2.7 & 72.5\% & 3.8 \\
        \midrule
        
        HERO (L1 only) w/o Replan  & 73.3\% & 4.0 & 50.0\% & 2.5 & 70.0\% & 2.7 & 70.0\% & 2.7 & 66.0\% & 3.0 \\
        HERO (L2 only) w/o Replan  & 76.7\% & \textbf{2.2} & 43.3\% & \textbf{1.3} & 53.3\% & \textbf{1.4} & 46.7\% & \textbf{1.2} & 55.0\% & \textbf{1.5} \\
        HERO (L3 only) w/o Replan  & 66.7\% & 3.2 & 46.7\% & 2.8 & 60.0\% & 1.9 & 66.7\% & 1.8 & 60.0\% & 2.4 \\

    \bottomrule 
    \end{tabular}

    \caption{Performance comparison of different HERO configurations.}
    \label{tab:performance}
\end{table*}

 \textbf{Capability Evolution} Figure \ref{fig:evolution} shows that as data collection progresses, the accumulated experiences gradually expand subtask coverage, enabling exemplar reuse to be invoked more frequently and further accelerating experience acquisition. Additionally, the reflexive policy continuously improves performance as more collected experiences are utilized for training. These results demonstrate the effective transition from autonomous experience acquisition and accumulation to capability evolution.

\subsection{Adaptive Task Execution}

We answer \textbf{Q2} in this section by evaluating the effectiveness of HERO's R$\rightarrow$E$\rightarrow$H capability execution strategy. We conducted \textbf{30} trials for each task, and evaluate framework performance using two metrics: (1) task success rate (SR), which measures the percentage of successfully completed tasks; and (2) successful task completion time (TS), which measures the average execution time over successful tasks.

\textbf{Capability Composition.} As illustrated in Table \ref{tab:performance}, individual capability variants reveal different mechanisms for exploiting accumulated experiences. L1 performs online reasoning, providing strong generalization across scenarios (76.0\% SR). However, when reliable experience patterns are already available, reasoning from scratch with VLMs introduce unnecessary uncertainty. Combining L1 with L3 improves performance to 82.7\%, showing that validated internalized experiences can effectively complement reasoning.

L2 enables efficient adaptation by directly reusing previously collected trajectories (2.8 min TS). However, transferred experiences may not always fully match the current execution context, limiting the effectiveness of direct reuse. 
Incorporating L1 provides additional reasoning when direct transfer is insufficient, improving L1+L2 over L2-only execution (73.3\% vs. 66.7\% SR). 
Meanwhile, L2+L3 improves L3-only execution (72.5\% vs. 70.0\%), suggesting that explicit exemplar reuse can complement behaviors that are not yet fully captured by the learned policy.

By integrating all three capabilities, HERO achieves the highest average success rate (86.0\%) across four tasks while maintaining an acceptable successful task completion time (3.9 min), demonstrating a favorable balance between robustness and efficiency. This demonstrates that robust manipulation benefits from flexibly exploiting accumulated experiences in different forms, where reasoning, exemplar reuse, and reflexive execution collectively bridge experience acquisition and reliable task execution.

\textbf{Reflexive Execution Analysis. }  While L1 and L2 generate behaviors based on the initial observations, L3 performs closed-loop execution by continuously adapting actions according to updated observations. Under randomly introduced object displacements during package picking and block stacking, L3 maintains an average subtask success rate of 67.2\%, while L1 and L2 execution cannot reliably handle such perturbations. This demonstrates the necessity of closed-loop reflexive execution for robust manipulation.

\textbf{Effect of Replanning. } Beyond capability selection, we further investigate the role of execution-time replanning, presented in Table \ref{tab:performance}.  As shown, removing the replanning mechanism  reduces execution time by avoiding feedback-based updates, but consistently degrades task success rates across all capability variants. L1-only execution decreases from 76.0\% to 66.0\% success rate without replanning, while L2-only and L3-only executions drop from 66.7\% to 55.0\% and from 70.0\% to 60.0\%, respectively. This indicates that continuously monitoring execution outcomes and updating dispatched subtasks is essential for handling unexpected failures and maintaining robust long-horizon manipulation.

\subsection{System Analysis}

\begin{table}[t]
\centering
\fontsize{9pt}{11pt}\selectfont

\begin{tabular}{l c c c}

\toprule
\textbf{Component} & \textbf{Frequency} & \textbf{Per-call (s)} & \textbf{Cost (s)} \\
\midrule

\multicolumn{4}{l}{\textbf{Orchestrator}} \\
\quad Planning & 0.29 & 7.95 & 2.33 \\
\quad Monitoring & 1.00 & 7.62 & 7.62 \\
\quad Replanning & 1.00 & 9.53 & 9.53 \\
\quad \textbf{Total} & & & \textbf{19.48} \\

\midrule

\multicolumn{4}{l}{\textbf{L1}} \\
\quad Skill Planning & 1.00 & 7.67 & 7.67 \\
\quad DA3 & 1.86 & 0.29 & 0.54 \\
\quad Keypoint Annotation & 1.86 & 11.31 & 21.04 \\
\quad Execution & 1.86 & 9.31 & 17.32 \\
\quad \textbf{Total} & & & \textbf{46.57} \\

\midrule

\multicolumn{4}{l}{\textbf{L2}} \\
\quad Retrieval & 1.00 & 0.22 & 0.22 \\
\quad DA3 & 1.86 & 0.28 & 0.52 \\
\quad SAM3 & 1.86 & 0.35 & 0.65 \\
\quad Motion Transfer & 1.86 & 1.35 & 2.51 \\
\quad Execution & 1.00 & 17.06 & 17.06 \\
\quad \textbf{Total} & & & \textbf{20.96} \\

\midrule

\multicolumn{4}{l}{\textbf{L3}} \\
\quad Execution & 1.00 & 37.90 & 37.90 \\
\quad \textbf{Total} & & & \textbf{37.90} \\

\bottomrule
\end{tabular}

\caption{Latency breakdown normalized to per-subtask execution. Frequency denotes the average number of module invocations per subtask.}
\label{tab:latency}
\end{table}


We now answer \textbf{Q3} by analyzing HERO's runtime behavior and system characteristics during deployment.

\textbf{Capability Utilization.} During deployment, HERO primarily relies on evolved reflexive policies while retaining L1 and L2 to handle subtasks through reasoning and experience transfer when needed. Among all successful subtasks, 86.7\% are directly completed by L3, 6.5\% require fallback to L2, while the remaining 6.8\% rely on L1. This distribution reflects the progressive consolidation of experience into executable policies: reflexive policies handle the majority of established behaviors, while other capabilities remain available to address cases beyond the current policy coverage.

\textbf{Efficiency Analysis.} As shown in Table \ref{tab:latency}, L1 has the highest latency (46.57s), mainly dominated by VLM-based reasoning and keypoint annotation for zero-shot generalization. By reusing previously collected experiences, L2 reduces the cost to 20.96s, while L3 visuomotor policy provides stable closed-loop execution (37.90s). These results reveal the computational trade-offs among capability layers: L1 improves generalization, L2 enhances efficiency through experience reuse, and L3 enables reliable real-time control.

\begin{figure}[t]
    \centering
    \includegraphics[width=1.0\linewidth]{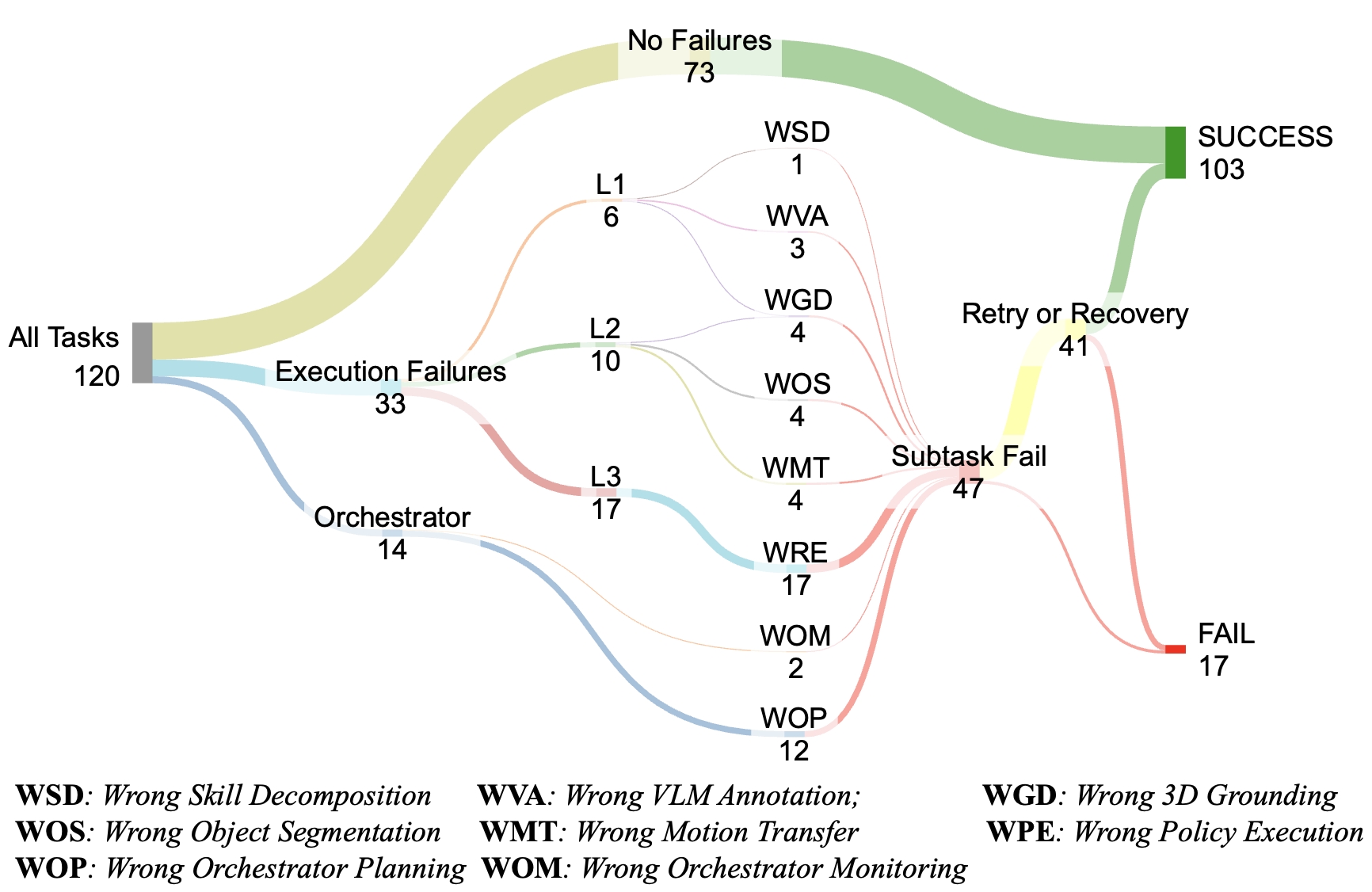}
    \caption{Failure Statistics in Task Execution}
    \label{fig:sankey}
\end{figure}

\textbf{Failure Analysis.} We analyze failure modes over all 120 task executions. As shown in Figure \ref{fig:sankey}, 73 tasks are completed without any failure, while failures in the remaining executions result from either execution-level errors (33 cases) and orchestrator-level errors (14 cases). Through closed-loop recovery, 41 intermediate failures trigger retry or recovery procedures, recovering 30 cases and resulting in 103 successful task completions overall.

 Among execution-level failures, L3 accounts for the largest portion (17/33), all attributed to wrong policy execution (WPE). This is consistent with its high invocation frequency during deployment, where L3 completes the majority of the subtasks. L1-related failures mainly originate from zero-shot semantic grounding, including wrong VLM annotation (WVA, 3). L2-related failures are mainly associated with experience transfer, including wrong object segmentation (WOS, 4) and wrong motion transfer (WMT, 4). Wrong 3D grounding (WGD, 4) affects both L1 reasoning and L2 experience reuse, highlighting the importance of accurate geometric grounding for reliable execution.

At the orchestrator level, 12 failures are caused by wrong orchestrator planning (WOP), while only 2 are attributed to wrong monitoring (WOM). This suggests that execution assessment is generally reliable, whereas generating optimal subtask plans for complex manipulation remains challenging.

Overall, these results demonstrate that HERO's closed-loop orchestration effectively recovers from most intermediate failures, while the remaining failures primarily reflect the inherent challenges of perception grounding, experience transfer, policy generalization, and task-level planning.
\section{Limitations and Future Work}

In this work, we propose HERO, an agentic framework that bootstraps and consolidates robotic capabilities from zero human demonstrations. HERO organizes robotic intelligence into three capability levels: heuristic reasoning for zero-shot experience acquisition, exemplar reuse for one-shot experience fast expansion, and reflexive execution for learned closed-loop control. Through hierarchical orchestration, HERO unifies autonomous data collection and task execution, enabling continuous capability evolution from general reasoning to increasingly embodied capabilities. Extensive experiments demonstrate its effectiveness in both autonomous capability evolution and adaptive task execution. 

Despite these advances, several challenges remain. First, VLM-based reasoning introduces noticeable inference latency and potential visual grounding errors, which may affect zero-shot experience acquisition. Second, HERO relies on accurate depth estimation and pointcloud reconstruction, making performance sensitive to perception noise and occlusions. Finally, the current framework bootstraps capability evolution from a predefined primitive skill space, whose design still requires human expertise and limits the ability to autonomously discover new skill abstractions. Future work will explore more efficient reasoning, robust perception, and open-ended skill discovery for further capability evolution.

\bibliography{aaai2027}
\clearpage
\twocolumn[
  \begin{center}
    {\Large\bfseries Supplementary Material}
    \vspace{2em}
  \end{center}
]

\section{Heuristic Bootstrapper}


\subsection{Annotation Post-Processing}


After the VLM predicts semantic keypoints and the corresponding 3D end-effector poses are obtained through geometric lifting, we further apply several geometric post-processing steps to improve execution consistency.

\textbf{End-Effector Offset Compensation. } The 3D poses generated from semantic annotations represent the desired interaction points on the object surface, while the robot controller executes motions through the tool center point (TCP). Therefore, directly executing these poses may cause the gripper body to collide with the object surface. To account for the physical offset between the TCP and the gripper tip, we translate the generated pose along the opposite direction of the TCP z-axis by the gripper length:
\begin{equation}
    t^{*}_{ee} = t_{ee} - 0.15 z_{ee}
\end{equation}
where $z_{ee}$ denotes the TCP z-axis direction and $0.15$ m is the measured distance between the wrist TCP and the gripper tip.

\textbf{Depth Ambiguity Correction. } For path-following skills (Push and Pull), directly lifting annotated 2D keypoints using the corresponding depth values may introduce geometric errors. Since semantic keypoints are defined in image space, the projected pixel may occasionally correspond to an incorrect depth surface due to occlusion or layered object structures. 

To resolve such ambiguities, we apply a geometry-based ray-plane intersection correction. For each skill, the keypoint with reliable depth is used to define the target interaction plane, while the ambiguous keypoint is reconstructed on this plane. Specifically, given a pixel location $(u,v)$, we construct its 3D viewing ray:

\begin{equation}
    P(t) = C+td(u,v)
\end{equation}
where $C$ is the camera center and $d(u,v)$ is the corresponding viewing direction in the world frame. The corrected 3D position is obtained by intersecting this ray with the target depth plane defined by the reliable keypoint. For Push, the push end point is used to correct the push start point; for Pull, the pull start point is used to correct the pull end point. This correction is activated when the depth difference between the two keypoints exceeds 5 cm, indicating that they likely correspond to different physical surfaces.

\textbf{Push Direction Disambiguation. } For Push skill, we align the TCP x-axis with the intended pushing direction to maximize contact area. Specifically, if the projection between the predicted TCP x-axis and the robot base x-axis satisfies:
\begin{equation}
    x_{ee} \cdot x_{base} > 0
\end{equation}
the current TCP x-axis direction is assigned as the pushing direction. Otherwise, the TCP x-axis is flipped to its opposite direction. This adjustment ensures consistent orientation generation for pushing motions and avoids unnecessary inverse kinematics failures caused by unfavorable end-effector configurations.

\subsection{Skill Constraints}
Although primitive skills provide executable manipulation targets, naively sequencing them can lead to collisions or unstable contact. We address this with a small set of geometric constraints that are automatically composed at runtime based on each skill's interaction type and the transition context.

\textbf{Skill-Local Pre-Conditions.}   Each skill defines geometric pre-conditions according to its interaction geometry. The corresponding approach direction $v_{\text{app}}$ is determined by the manipulation objective: for grasp and place, it follows the TCP $z$-axis; for push and pull, it follows the intended interaction direction. Before executing the target $p_{\text{target}} $, the skill automatically generates one or more pre-approach poses by offsetting from the target configuration:
\begin{equation}
    p_{\text{pre}} = p_{\text{target}} - d \cdot v_{\text{app}}
\end{equation}
where $d$ is set to 0.15~m. This ensures the end-effector approaches the target from a consistent, safe direction. 

\textbf{Cross-Skill Transition. } When switching between skills of different types, a retract step is prepended to the incoming skill's pre-condition sequence. The retract shares the same form:
\begin{equation}
    p_{\text{retract}} = p_{\text{ee}} - d \cdot v_{\text{app}}^{\text{prev}}
\end{equation}
where $p_{\text{ee}}$ is the current end-effector pose and $v_{\text{app}}^{\text{prev}}$ is inherited from the preceding skill's interaction direction. This lifts the arm clear of the previous contact state before re-approaching the new target.



\section{Exemplar Accelerator}

\subsection{Demonstration Retrieval}

\textbf{Text matching.} Demonstrations are indexed by their natural-language task descriptions. Retrieval uses ordered Longest Common Subsequence (LCS) \cite{LCS} similarity with an F1-style score: $s(a,b)=2\cdot\text{LCS}(a,b)/(|a|+|b|)$. Stop words (\emph{the, a, on, in, to, from, and, or, it, into, then, put, with, of}) are removed before matching. Only demonstrations with a perfect matching score ($s=1.0$) are considered candidates for exemplar reuse. After text filtering, demonstrations whose stored object point clouds contain fewer than 1000 points are excluded, as sparse reconstructions may provide unreliable geometric correspondence for motion transfer.

\textbf{Quality scoring.} Considering that successful demonstrations may differ in transferability, each demonstration maintains a quality score $q \in (0.0,2.0]$ (initialized to 1.0). Successful reuse increases the score by $+0.1$, while failed transfers decrease it by $-0.3$. Candidate demonstrations are ranked according to their quality scores, with higher-transferability exemplars prioritized for reuse. Demonstrations with $q\leq0.0$ are removed from the retrieval library.



\subsection{Motion Transfer}

Given a retrieved exemplar, L2 transfers the demonstrated end-effector trajectory to the current scene by estimating the relative transformation between the reference object and the current object. The transformation estimation follows a coarse-to-fine registration strategy.

\textbf{Coarse Initialization. } We first obtain an initial object transformation from the stored demonstration object point cloud and the current object point cloud. The translation component is initialized by aligning the two object centroids, while a rotation regressor from MT3 \cite{MT3} predicts the initial rotational transformation. This provides a coarse estimate:
\begin{equation}
    T_{init} = (R_{init}, T_{init})
\end{equation}

\textbf{Fine Registration. } Starting from the coarse initialization, we further refine the transformation using Generalized ICP (GICP) \cite{GICP}. To improve robustness against local minima, multiple GICP optimizations are performed with translation perturbations of $\pm 2$ cm around the initial estimate. The transformation with the lowest inlier RMSE is selected as the refined result. GICP uses a maximum correspondence distance of 0.10 m, 50 iterations per restart, and a total timeout of 1.0 s. Registration results with fewer than 500 valid correspondences are rejected as unreliable. Additionally, if the refined yaw angle deviates from the initial estimate by more than 15$^\circ$, the GICP result is considered unreliable and the initial transformation is retained.

\textbf{4-DOF Inductive Bias. } Although full 6-DoF registration provides the best geometric alignment in ideal conditions, noisy point clouds may introduce unnecessary roll and pitch variations that are undesirable for real robot trajectory transfer. Following the inductive bias adopted in MT3 \cite{MT3}, we constrain the final transformation to 4-DoF, retaining only translation and yaw rotation while removing roll and pitch components.

Specifically, given the original 6-DoF transformation:
\begin{equation}
    P_{current} = R_{6dof}P_{demo} + t_{6dof}
\end{equation}
where $P_{demo}$ and $P_{current}$ are the demo and current end-effector poses, respectively.

We construct the constrained transformation:
\begin{equation}
    P_{current} = R_{4dof}P_{demo} + t_{4dof}
\end{equation}
where $R_{4dof}$ only contains the yaw rotation.

To preserve the original end-effector reference position, the translation is compensated as:
\begin{equation}
    t_{4dof} = R_{6dof}P_{demo} + t_{6dof} - R_{4dof}P_{demo} 
\end{equation}

The resulting 4-DoF transformation is finally applied to warp the demonstrated end-effector trajectory into the current scene.




\section{Training and Inference}

\subsection{Model Initialization and Hyperparameters}

With regards to the L3 implementation, we fine-tune the publicly released $\pi_{0.5}$-base checkpoint \cite{Pi0.5} for all experiments. Training hyperparameters follow the official fine-tuning recipe unless noted otherwise:

\begin{itemize}
  \item \textbf{Seed}: 0 (default).
  \item \textbf{Optimizer}: AdamW with $\beta_1{=}0.9$, $\beta_2{=}0.999$, weight decay $10^{-4}$, gradient clipping at norm 1.0.
  \item \textbf{Learning rate schedule}: cosine decay from $5\times10^{-5}$ to $10^{-6}$, with 50 linear warmup steps.
  \item \textbf{Batch size}: 32 per 2 GPUs 
  \item \textbf{EMA}: exponential moving average with decay 0.99.
  \item \textbf{Training steps}: 10K steps for each task.
  \item \textbf{Action representation}: $\pi_{0.5}$ predicts 32-dimensional actions, while only the first 8 dimensions $[j_0,\dots,j_6,g]$ are used for FR3 joint-level control.
\end{itemize}

\subsection{Action Chunking and Execution Horizon}

The model predicts an action chunk of 50 steps, which corresponds to approximately 3.3 seconds of future trajectory at 15 Hz. During closed-loop inference, we execute the first 30 predicted actions before re-inferring, following the receding horizon control paradigm. 


\subsection{Language Injection into the Action Expert}

We empirically observe that the standard $\pi_{0.5}$ architecture, which fuses language tokens with image tokens in a shared PaliGemma \cite{paligemma} prefix before cross-attending to the Action Expert suffix, produces policies that are insensitive to color specifications in language prompts (e.g., ``pick up the red block'' vs.\ ``pick up the blue block''). We hypothesize that the language signal is diluted by the dominant visual features: $\sim$20 text tokens compete for attention with $\sim$512 image tokens across the 18 shared transformer layers.

To address this, we introduce a lightweight \textbf{language injection} pathway. A linear projection layer $\mathbf{W}_{\text{lang}} \in \mathbb{R}^{2048 \times 1024}$ (approximately 2.1M parameters) maps the mean-pooled PaliGemma text embeddings directly into the Action Expert's token space. The resulting language vector is broadcast-added to all 50 action tokens before they enter the shared transformer, creating a direct shortcut that bypasses the image-dominated cross-attention mechanism:

\begin{equation}
\mathbf{h}_{\text{lang}} = \text{Swish}(\mathbf{W}_{\text{lang}} \cdot \text{mean}(\mathbf{E}_{\text{text}}, \text{axis}{=}1))
\end{equation}

\begin{equation}
\quad \mathbf{A}'_t = \mathbf{A}_t + \mathbf{h}_{\text{lang}}
\end{equation}
where $\mathbf{E}_{\text{text}} \in \mathbb{R}^{T \times 2048}$ denotes the PaliGemma text token embeddings and $\mathbf{A}_t \in \mathbb{R}^{50 \times 1024}$ denotes the action tokens. This layer is randomly initialized and trained from scratch alongside the fine-tuned model weights. 

\subsection{Data Filtering and Gripper Processing}

Raw L1 collected demonstrations are converted to the LeRobot\cite{lerobot} format for training. Stationary frames where the robot is idle during VLM reasoning are removed: a frame is dropped when $\lVert\mathbf{j}_{t+1} - \mathbf{j}_t \rVert_\infty < 0.001$, where $\mathbf{j}_t \in \mathbb{R}^{7}$ stands for the joint position of robot at frame t. Gripper values are binarized at a threshold of 0.5 and temporally debounced with a 15-frame window to suppress rapid open-close cycles at grasp and release points. Task descriptions are kept as raw natural language without template simplification. 

\subsection{Dataset Statistics}

\begin{table}[h]
\centering
\begin{tabular}{lrl}
\toprule
Task & Episodes  \\
\midrule
Package Picking & 180  \\
Block Stacking   & 163  \\
Drawer Search & 158  \\
Cuboid Discovery & 163  \\
\bottomrule
\end{tabular}
\caption{Task datasets used for $\pi_{0.5}$ fine-tuning. All datasets use raw natural-language prompts.}
\label{tab:datasets}
\end{table}

\subsection{Closed-Loop Inference and Termination}

Inference follows the same format as training: observations are captured from the robot at 15 Hz (front and wrist RGB cameras, joint positions and gripper state). Convergence is detected when $\lVert\mathbf{j}_{t+1} - \mathbf{j}_t \rVert_\infty < 0.02$ for 8 consecutive inference steps. A safety limit of 25 inference rounds prevents indefinite execution. Inference is performed on a single NVIDIA A100 GPU.

\end{document}